\documentclass{article} 
\usepackage{iclr2025_conference,times}

\usepackage{microtype}
\usepackage{pgfplots}
\pgfplotsset{compat=1.17}
\usepackage{graphicx}
\usepackage{subfigure}
\usepackage{booktabs} 
\usepackage{algorithm,algpseudocode}
\usepackage{amsmath,amssymb,mathtools,amsthm}
\usepackage{tikz}
\usetikzlibrary{shapes.geometric,positioning}
\usepackage{textcomp}
\usepackage{hyperref}
\usepackage[capitalize,noabbrev]{cleveref}
\PassOptionsToPackage{hyphens}{url}

\theoremstyle{plain}

\theoremstyle{definition}

\theoremstyle{remark}

\usepackage{fancyvrb}
\usepackage{fvextra}
\usepackage{microtype} 

\fvset{
  breaklines=true, 
  breakanywhere=true, 
  breakautoindent=true, 
  fontsize=\small,
  frame=single, 
  numbers=left 
}

\iclrfinalcopy

\title{MINDSTORES: Memory-Informed Neural Decision Synthesis for Task-Oriented Reinforcement in Embodied Systems}

\author{Anirudh Chari$^{1*}$ \quad Suraj Reddy$^{1*}$ \quad Aditya Tiwari$^{2}$ \quad Richard Lian$^{1}$ \quad Brian Zhou$^{3}$ \\
$^{1}$Massachusetts Institute of Technology, Cambridge, MA \\ $^{2}$Illinois Mathematics and Science Academy, Aurora, IL \\ $^{3}$Harvard University, Cambridge, MA \\
\texttt{\{anichari,surajrdy,rlian\}@mit.edu}; \texttt{atiwari@imsa.edu}, \\ \texttt{brianzhou@college.harvard.edu}
\\ * = Equal Contribution
}

\begin{document}

\maketitle

\begin{abstract}
While large language models (LLMs) have shown promising capabilities as zero-shot planners for embodied agents, their inability to learn from experience and build persistent mental models limits their robustness in complex open-world environments like Minecraft. We introduce MINDSTORES, an experience-augmented planning framework that enables embodied agents to build and leverage \textit{mental models} through natural interaction with their environment. Drawing inspiration from how humans construct and refine cognitive mental models, our approach extends existing zero-shot LLM planning by maintaining a database of past experiences that informs future planning iterations. The key innovation is representing accumulated experiences as natural language embeddings of (state, task, plan, outcome) tuples, which can then be efficiently retrieved and reasoned over by an LLM planner to generate insights and guide plan refinement for novel states and tasks. Through extensive experiments in the MineDojo environment, a simulation environment for agents in Minecraft that provides low-level controls for Minecraft, we find that MINDSTORES learns and applies its knowledge significantly better than existing memory-based LLM planners while maintaining the flexibility and generalization benefits of zero-shot approaches, representing an important step toward more capable embodied AI systems that can learn continuously through natural experience.
\end{abstract}

\section{Introduction}

Recent advances in large language models (LLMs) have demonstrated enhanced capabilities in reasoning \citep{plaat_reasoning_2024, huang_towards_2023}, planning \citep{sel_llms_2025}, and decision-making \citep{huang_making_2024} through methods that strengthen analytical depth. Among the numerous domains of active innovation, the success of AI agents serves as a critical benchmark for assessing our progress toward generally capable artificial intelligence \citep{brown_language_2020}.

Building \textit{embodied} agents---AI systems with physical form---that learn continuously from real-world interactions through persistent memory and adaptive reasoning remains a fundamental challenge in the future of artificial intelligence. Classical approaches, such as reinforcement learning \citep{dulac-arnold2021challenges} and symbolic planning \citep{10694733}, struggle with scalability, irreversible errors, and rigid assumptions in complex environments.

A promising paradigm for such agents leverages LLMs as high-level planners \citep{jeurissen2024playingnethackllmspotential}: the LLM decomposes abstract goals into step-by-step plans (e.g., “mine wood $\rightarrow$ craft tools $\rightarrow$ smelt iron”), while a low-level controller translates these plans into environment-specific actions (e.g., movement, object interaction). This “brain and body” architecture capitalizes on the LLM’s capacity for structured reasoning while grounding its outputs in the dynamics of the physical world—a critical capability for real-world applications like robotic manipulation \citep{shentu2024llmsactionslatentcodes,bhat2024groundingllmsrobottask,WANG2024}, autonomous navigation \citep{zawalski2024roboticcontrolembodiedchainofthought}, and adaptive disaster response.

\begin{figure}[ht]
\centering
\scalebox{0.75}{
\begin{tikzpicture}[
    node distance=2cm,
    startstop/.style={rectangle, rounded corners, minimum width=3cm, minimum height=1cm, text centered, draw=black, fill=red!40},
    process/.style={rectangle, rounded corners, minimum width=3cm, minimum height=1cm, text centered, draw=black, fill=gray!20},
    database/.style={rectangle, rounded corners, scale=0.8, fill=orange!10, minimum height=0.7cm, minimum width=12cm},
    plan/.style={rectangle, rounded corners, scale=0.8, fill=green!10, minimum height=0.7cm, minimum width=12cm},
    execution/.style={rectangle, rounded corners, scale=0.8, fill=purple!10, minimum height=0.7cm, minimum width=12cm},
    arrow/.style={thick,->,>=stealth}
]
    \node (input) [startstop] {Input};
    \node (review) [process, below of=input] {Review Experience};
    \node (plan) [process, below of=review] {Plan Generation};
    \node (predict) [process, below of= plan] { Predict Outcome };
    \node (execute) [process, below of=predict] {Execute};
    \node (record) [startstop, below of=execute] {Record};

    \draw [arrow] (input) -- (review);
    \draw [arrow] (review) -- (plan);
    \draw [arrow] (plan) -- (predict);
    \draw [arrow] (predict) -- (execute);
    \draw [arrow] (execute) -- (record);
    
    \draw [arrow] (predict.east) --++ (1,0) |- (review.east)
        node[midway, right, rotate=-90, yshift=0.25cm, xshift=0.3cm] {Failure Management};

    \draw [arrow] (record.west) --++ (-1,0) |- (input.west)
        node[midway, left, rotate=90, yshift=0.25cm, xshift=-3cm] {Next Iteration};
    
    \fill[gray!10, rounded corners] (4, 0) rectangle (14.75, -8);
    \node [anchor=west] at (4.5, -0.5) (prompt) {Task: Mine Iron};
    \node [database, anchor=west] at (4.5, -1.2) {Review: Previous experiences for mining needs a pickaxe!};
    \node [plan, anchor=west] at (4.5, -1.9) {Plan: Mine \includegraphics[height=0.5cm]{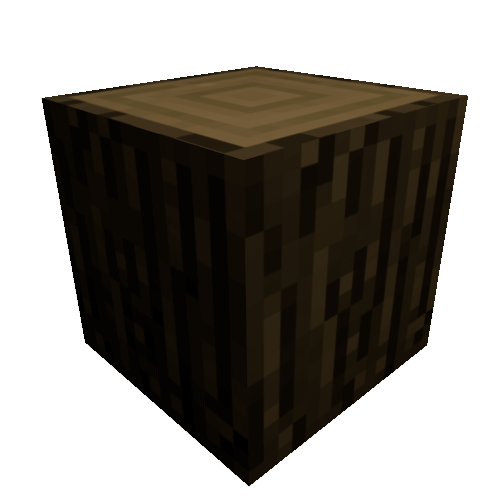}, craft \includegraphics[height=0.5cm]{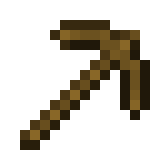}, mine \includegraphics[height=0.5cm]{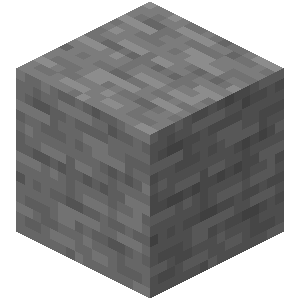}, craft \includegraphics[height=0.5cm]{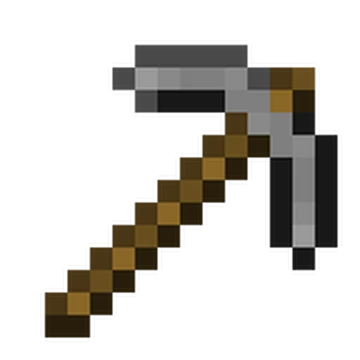}, mine \includegraphics[height=0.5cm]{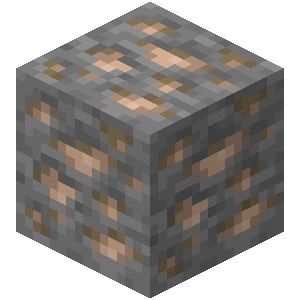}};
    \node [database, anchor=west] at (4.5, -2.6) {Predict: Probable failure, might die from hunger while finding iron ore.};
    \node [database, anchor=west] at (4.5, -3.3) {Review: Find food before searching for iron ore.};
    \node [plan, anchor=west] at (4.5, -4) {Plan: Mine \includegraphics[height=0.5cm]{wood.png}, craft \includegraphics[height=0.5cm]{woodenpick.png}, mine \includegraphics[height=0.5cm]{stone.png}, craft \includegraphics[height=0.5cm]{stonepic.png}, craft \includegraphics[height=0.5cm]{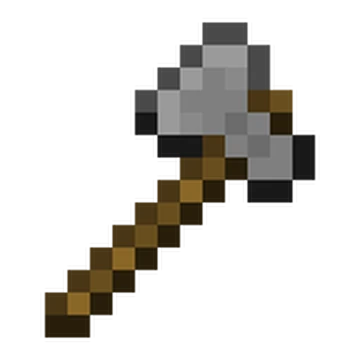}, hunt \includegraphics[height=0.5cm]{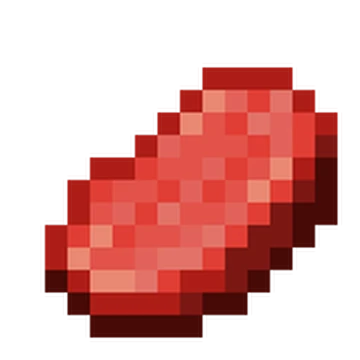}, mine \includegraphics[height=0.5cm]{iron.png}};
    \node [database, anchor=west] at (4.5, -4.7) {Predict: Probable success!};
    \node [execution, anchor=west] at (4.5, -6) {
        \begin{tabular}{c}
        Execution: \\
        \includegraphics[width=2.8cm]{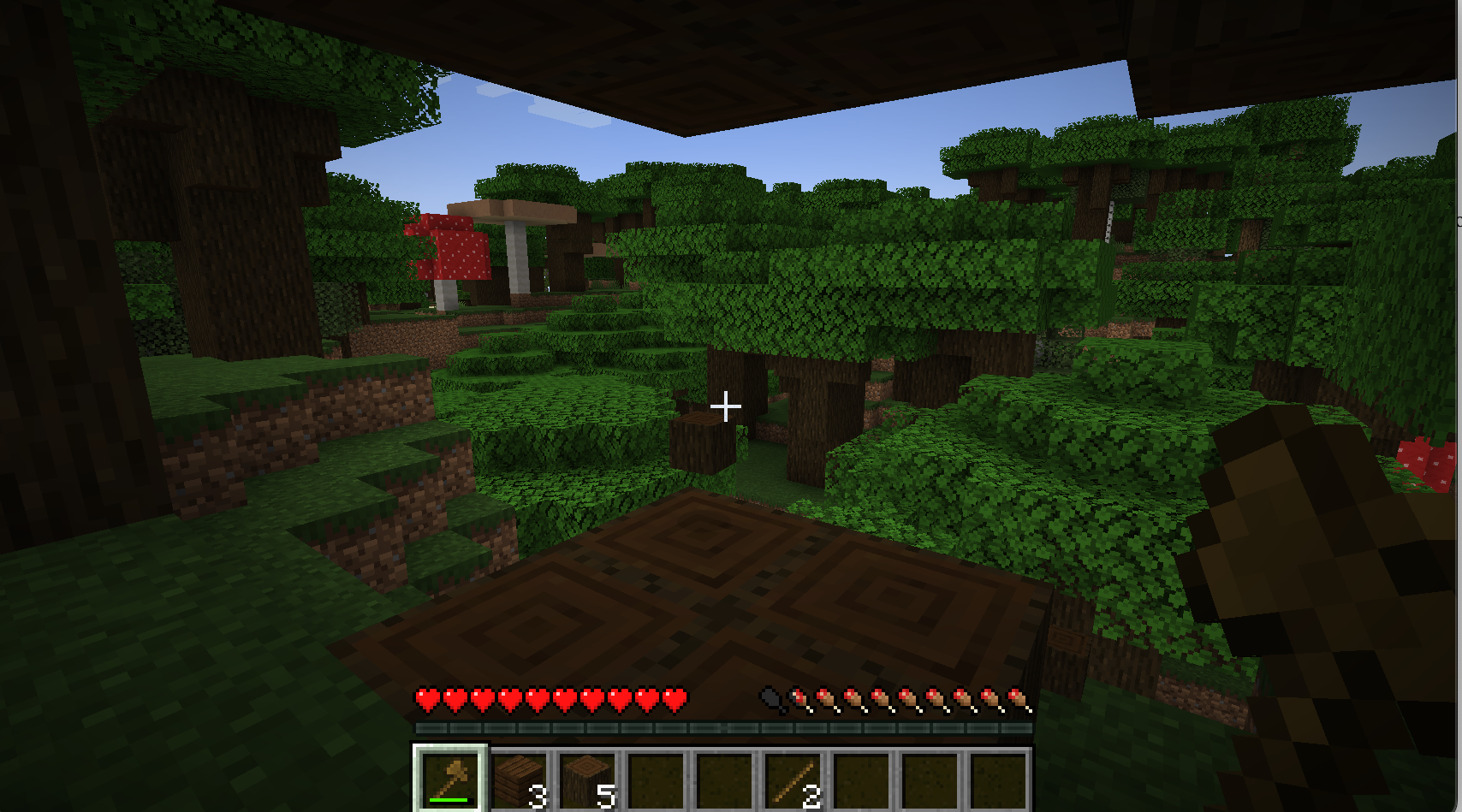} \quad \includegraphics[width=2cm]{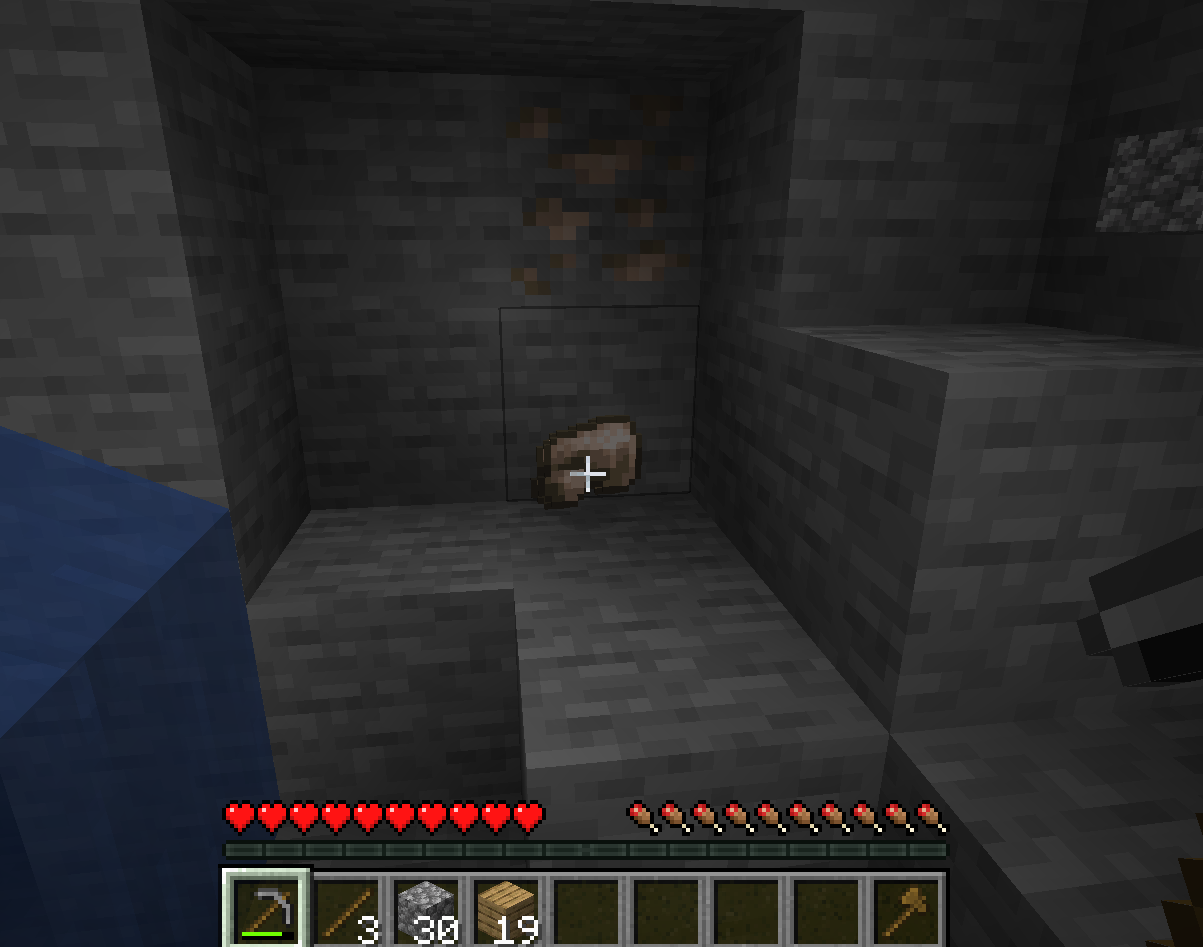}
        \end{tabular}
    };
    \node [database, anchor=west] at (4.5, -7.4) {Store relevant information back into database to refer to later.};
\end{tikzpicture}
}
\caption{Overview of the MINDSTORES planning architecture. The left shows the iterative experiential learning pipeline leveraging the experience database. Database-related methods are in orange, planning steps are in green, and Minecraft steps are in red. The right shows an example applying this pipeline to an example task in Minecraft.}
\label{fig:overview}
\end{figure}

While recent LLM-based agents show promise in generating action plans for embodied tasks, many lack  \textit{experiential} learning, i.e., the ability to apply insights from past experiences to planning for future tasks. Unlike humans—who build mental models to generalize insights, avoid errors, and reason counterfactually (e.g., “Crafting a stone pickaxe first would enable iron mining”)—existing agents cannot synthesize persistent representations of past interactions. This gap hinders their ability to tackle long-horizon tasks in open worlds like Minecraft, where success requires inferring objectives, recovering from failures, and transferring insights across scenarios.

Minecraft exemplifies these challenges: agents must explore procedural terrains, infer task dependencies (e.g., stone tools before iron mining), and adapt to unforeseen challenges. Current LLM planners, namely zero-shot architectures like DEPS \citep{wang_describe_2024}, exhibit critical flaws: (1) they lack persistent mental models, causing repetitive errors (e.g., using wooden pickaxes for iron mining); and (2) they underutilize LLMs’ reasoning to synthesize experiential insights, producing brittle plans.

To address these limitations, we propose MINDSTORES, a framework that leverages LLMs to construct dynamic mental models—internal representations guiding reasoning and decision-making, inspired by human cognition. Just as humans build simplified models of reality to anticipate events and solve problems, our approach equips agents to actively interpret experiences through structured reasoning. By analyzing failures (e.g., “Wooden pickaxes break mining iron”), inferring causal rules (e.g., “Stone tools are prerequisites”), and predicting outcomes, the LLM transforms raw interaction data into adaptive principles.

MINDSTORES augments planners with an experience database storing natural language tuples (state, task, plan, outcome) and operates cyclically: observe, retrieve relevant experiences, synthesize context-aware plans, act, and log outcomes. This closed-loop process enables semantic analysis of memories, iterative strategy refinement, and outcome prediction, bridging the gap between static planning and experiential learning while grounding agent reasoning in human-like cognitive foundations.

Hence, our key contributions are as follows:
\begin{itemize}
    \item A cognitive-inspired formulation of artificial mental models to enable natural-language memory accumulation and transfer learning.
    \item \textbf{MINDSTORES}, a novel open-world LLM planner leveraging the above formulation to develop lifelong learning embodied agents.
    \item Extensive evaluation of MINDSTORES in Minecraft, demonstrating a \textbf{9.4\%} mean improvement in open-world planning tasks over existing methods.
\end{itemize}

In the remainder of this paper, we detail the theoretical foundations of mental models in Section~\ref{section:bg}, present the MINDSTORES architecture in Section~\ref{section:md}, and validate its performance through experiments in Sections~\ref{section:ep} and~\ref{section:ra}. Our findings underscore the critical role of memory-informed reasoning in developing lifelong learning agents for open-world environments.

\section{Background}
\label{section:bg}
\subsection{Open-World Planning for Embodied Agents}
Planning for embodied agents in open-world environments presents unique challenges due to the unbounded action space, long-horizon dependencies, and complex environmental dynamics. In environments like Minecraft, agents must reason about sequences of actions that may span dozens of steps, where early mistakes can render entire trajectories infeasible \citep{fan_minedojo_2022}. Traditional planning approaches that rely on explicit state representations and value functions struggle in such domains due to the combinatorial explosion of possible states and actions.

The key challenges in open-world planning stem from two main factors. First, the need for accurate multi-step reasoning due to long-term dependencies between actions presents a significant hurdle. Second, the requirement to consider the agent's current state and capabilities when ordering parallel sub-goals within a plan poses additional complexity. Consider the example of crafting a diamond pickaxe in Minecraft: the process requires first obtaining wood, then crafting planks and sticks, mining stone with a wooden pickaxe, crafting a stone pickaxe, mining iron ore, smelting iron ingots, and finally crafting the iron pickaxe---a sequence that can easily fail if any intermediate step is incorrectly executed or ordered.

\subsection{Zero-Shot LLM Planning with DEPS}
Recent work has shown that large language models can serve as effective zero-shot planners for embodied agents through their ability to decompose high-level tasks into sequences of executable actions \citep{huang_language_2022}. The DEPS (Describe, Explain, Plan and Select) framework leverages this capability through an iterative planning process that combines several key components \citep{wang_describe_2024}. The framework utilizes a descriptor that summarizes the current state and execution outcomes, an explainer that analyzes plan failures and suggests corrections, a planner that generates and refines action sequences, and a selector that ranks parallel candidate sub-goals based on estimated completion steps.

The key innovation of DEPS is its ability to improve plans through verbal feedback and explanation. When a plan fails, the descriptor summarizes the failure state, the explainer analyzes what went wrong, and the planner incorporates this feedback to generate an improved plan. This creates a form of zero-shot learning through natural language interaction.

However, DEPS and similar approaches maintain no persistent memory across episodes. Each new planning attempt starts fresh, unable to leverage insights gained from previous successes and failures in similar situations. This limitation motivates our work on experience-augmented planning.

\subsection{Mental Models}
Mental models are cognitive representations of how systems and environments work, enabling humans to understand, predict, and interact with the world around them. Originally proposed by \citet{craik_nature_1952}, mental models theory suggests that people construct small-scale internal models of reality that they use to reason, anticipate events, and guide behavior. These models are built through experience and observation, continuously updated as new information becomes available, and help reduce cognitive load by providing ready-made frameworks for understanding novel situations.

A key insight from psychological research on mental models is their role in transfer learning and generalization \citep{canini_modeling_nodate}. When faced with new scenarios, humans naturally draw upon their existing mental models to make informed decisions, even in previously unseen contexts. This ability to leverage past experiences through abstract representations is particularly relevant for embodied agents operating in open-world environments, where they must constantly adapt to novel situations while maintaining coherent, generalizable knowledge about environmental dynamics.

\section{Methods}
\label{section:md}
\begin{figure}
\includegraphics[width=\linewidth]{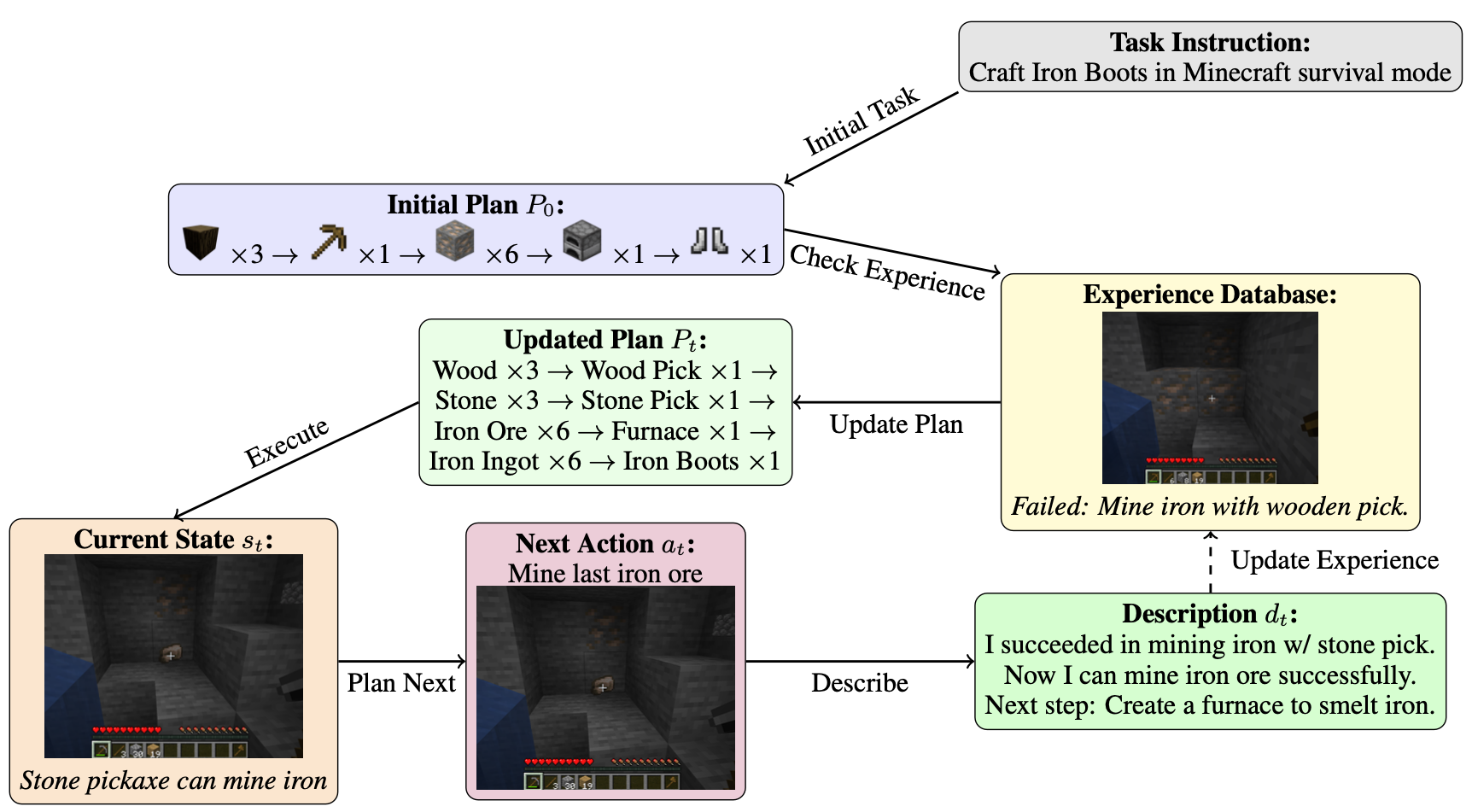}
\caption{Interactive planning process for crafting iron boots in Minecraft. The system initially plans to mine iron with a wooden pickaxe but learns from past experience that this will fail. It then updates the plan to include creating a stone pickaxe first, leading to successful iron ore mining.}
\end{figure}

\subsection{Overview}

We propose an experience-augmented planning framework that maintains a similar foundation to DEPS but advances by maintaining a persistent mental model of the environment through natural language experiences. Our approach integrates several key components into a cohesive system. The framework maintains a database $\mathcal{D}$ of experience tuples $(s, t, p, o)$ containing state descriptions $s$, tasks $t$, plans $p$, and outcomes $o$. This is complemented by a semantic retrieval system for finding relevant past experiences, an LLM planner that generates insights and plans informed by retrieved experiences, and a prediction mechanism that estimates plan outcomes before execution.

\subsection{Experience Database}

Each experience tuple $(s, t, p, o) \in \mathcal{D}$ consists of natural language paragraphs describing the environmental context. The state $s$ captures the environmental context and agent's condition. The task $t$ represents the high-level goal to be achieved. The plan $p$ contains the sequence of actions generated by the planner. Finally, the outcome $o$ describes the execution result and failure description if applicable.

For each component, we compute a dense vector embedding $e(x) \in \mathbb{R}^d$ using a pretrained sentence transformer, where $x$ represents any of $s$, $t$, $p$, or $o$. This allows efficient similarity-based retrieval using cosine distance:
\begin{equation}
\text{sim}(x_1, x_2) = \frac{e(x_1) \cdot e(x_2)}{\|e(x_1)\| \|e(x_2)\|}
\end{equation}

\subsection{Experience-Guided Planning}

Given a new state $s_t$ and task $t_t$, our algorithm proceeds through several stages. Initially, it retrieves the $k$ most similar past experiences based on state and task similarity:
\begin{equation}
\mathcal{N}_k(\mathcal{D}, s_t, t_t) = \text{top-k}_{(s,t,p,o) \in \mathcal{D}} \left[\sum_{x \in \{ s, t\} } \lambda_x \text{sim}(x, x_t)\right]
\end{equation}
The LLM is then prompted to analyze these experiences and generate insights about common failure modes to avoid, successful strategies to adapt, and environmental dynamics to consider. Following this analysis, it generates an initial plan $p_t$ conditioned on the state, task, experiences, and insights.

The system then predicts the likely outcome by retrieving similar past plans:
\begin{equation}
\mathcal{N}_k(s_t, t_t, p_t) = \text{top-k}_{(s,t,p,o) \in \mathcal{D}} \left[\sum_{x \in \{s, t, p \}} \lambda_x \text{sim}(x, x_t)\right]
\end{equation}
If predicted outcomes suggest likely failure, the system returns to the plan generation stage to revise the plan. Finally, it executes the plan and stores the new experience tuple in $\mathcal{D}$. The complete process is formalized in Algorithm~1.

\begin{algorithm}
\caption{Experience-Augmented Planning}
\begin{algorithmic}[1]
\Require State $s_t$, Task $t_t$, Database $\mathcal{D}$, LLM $M$, $k$ neighbors
\Ensure Plan $p_t$
\State $\mathcal{N}_k \gets \texttt{retrieve\_top\_k}(\mathcal{D}, s_t, t_t, k)$
\State $\text{insights} \gets M.\texttt{analyze\_experiences}(\mathcal{N}_k)$
\State $p_t \gets M.\texttt{generate\_plan}(s_t, t_t, \mathcal{N}_k, \texttt{insights})$
\While{true}
    \State $\text{similar\_plans} \gets \texttt{get\_similar\_plans}(\mathcal{D}, s_t, t_t, p_t)$
    \State $\text{pred\_outcome} \gets \texttt{analyze\_outcomes}(\text{similar\_plans})$
    \If{$\text{pred\_outcome}$ is success}
        \State \textbf{break}
    \EndIf
    \State $p_t \gets M.\texttt{revise\_plan}(p_t, \text{pred\_outcome})$
\EndWhile
\State $\text{outcome} \gets \texttt{execute\_plan}(p_t)$
\State $\mathcal{D}.\texttt{add}((s_t, t_t, p_t, \text{outcome}))$
\State \Return $p_t$
\end{algorithmic}
\end{algorithm}

\subsection{Design Justification}

Our approach incorporates several carefully considered design elements that work together to create an effective planning system. The use of natural language experiences, rather than vectors or symbolic representations, leverages the LLM's ability to perform flexible reasoning over arbitrary descriptions. The semantic retrieval system employs dense embeddings to enable efficient similarity search while capturing semantic relationships between experiences beyond exact matches. The two-stage retrieval process first retrieves experiences based on state/task similarity to inform plan generation, then retrieves similar plans to predict outcomes, allowing the planner to both learn from past experiences and validate new plans. Finally, the iterative refinement capability enables the planner to revise plans based on predicted outcomes before execution, reducing the cost of actual failures.

This design maintains the benefits of zero-shot LLM planning while enabling continual learning through natural experience.

\section{Experiments}
\label{section:ep}

\subsection{Experimental Setup}
We evaluate our experience-augmented planning approach in MineDojo using 8 tiers of task complexity (MT1--MT8) \citep{fan_minedojo_2022}. The observation space includes RGB view, GPS coordinates, and inventory state, with 42 discrete actions mapped from MineDojo's action space \citep{fan_minedojo_2022}. All experiments utilize the behavior cloning controller trained on human demonstrations, following similar methodology to DEPS and Voyager. Due to software version constraints, our implementation of the controller achieves lower baseline performance than the original DEPS controller. Therefore, we use our implementation of DEPS without the experience database as the primary baseline for fair comparison. Each task is evaluated over 30 trials with randomized initial states and a fixed random seed of 42.

Our experience database uses Sentence-BERT embeddings (768-dim) stored in FAISS for efficient search. Key parameters were determined through ablation studies:
\begin{itemize}
    \item Optimal \( k=5 \) neighbors (tested \( k=1,3,5,10,20 \))
    \item Weighted similarity: \(\lambda_s=0.4\) (state), \(\lambda_t=0.4\) (task), \(\lambda_p=0.2\) (plan)
\end{itemize}

For the complete agent algorithm and associated LLM prompts, see Appendix A, and for detailed implementation aspects including environment integration and neural component configurations, see Appendix B.

\subsection{Evaluation Tasks}
We evaluate on 53 Minecraft tasks grouped into 3 complexity tiers:
\begin{itemize}
    \item \textbf{Basic (MT1--MT2)}: Fundamental tasks (wood/stone tools, basic blocks)
    \item \textbf{Intermediate (MT3--MT5)}: Progressive tasks (food, mining, armor crafting)
    \item \textbf{Advanced (MT6--MT8)}: Complex tasks (iron tools, minecart, diamond)
\end{itemize}
Episode lengths range from 3,000 steps (Basic) to 12,000 steps (Challenge tasks).

For additional task details and performance statistics, see Appendix C.

\subsection{Baselines}
We compare MINDSTORES to the following existing state-of-the-art approaches:
\begin{itemize}
    \item \textbf{DEPS}: State-of-the-art zero-shot LLM planner \citep{wang_describe_2024}
    \item \textbf{Voyager}: Automated curriculum learning agent \citep{wang_voyager_2023}
    \item \textbf{Reflexion}: LLM planner with environmental feedback \citep{shinn_reflexion_2023}
\end{itemize}

\subsection{Ablations}
To analyze the function of each individual component of the MINDSTORES framework, we perform the following ablations:
\begin{itemize}
    \item \textbf{No Experience}: Remove retrieval component
    \item \textbf{Fixed \( k \) Values}: Test \( k=1,3,5,10,20 \) retrieval contexts
    \item \textbf{Single-Shot}: Disable iterative plan refinement (DEPS)
\end{itemize}

\subsection{Metrics}
To quantify each method's performance in open-world planning in the Minecraft environment, we measure:
\begin{itemize}
    \item \textbf{Success Rate}: Completion percentage across trials
    \item \textbf{Learning Efficiency}: Iterations required for skill mastery
    \item \textbf{Complexity Scaling}: Performance vs. task complexity tiers
    \item \textbf{Retrieval Impact}: Success rate vs. context size (\( k \))
    \item \textbf{Continuous Learning}: Effect of non-discrete experience database for each task progression
\end{itemize}

\section{Results and Analysis}
\label{section:ra}

Our experiments reveal significant performance differences between MINDSTORES and DEPS across task categories, highlighting key insights into their scalability and effectiveness.

\subsection{Performance Metrics}

\begin{figure}[ht]
\centering
\begin{minipage}[b]{0.48\textwidth}
\centering
\begin{tikzpicture}
\begin{axis}[
    width=\linewidth,
    height=0.7\linewidth,
    xlabel={Task Difficulty (MT)},
    ylabel={Success Rate (\%)},
    symbolic x coords={1,2,3,4,5,6,7,8},
    xtick=data,
    ymin=0,
    ymax=100,
    legend style={
        at={(0.5,1.25)},
        anchor=north,
        legend columns=2
    },
    ymajorgrids=true,
    grid style=dashed,
    bar width=0.15cm,
    ybar=0pt,
    enlarge x limits=0.15,
]
\addplot[red,fill=red!30] coordinates {
    (1,83.3) (2,83.7) (3,71.0) (4,55.6)
    (5,29.1) (6,20.5) (7,17.3) (8,0.0)
};
\addplot[blue,fill=blue!30] coordinates {
    (1,70.6) (2,77.0) (3,60.0) (4,42.8)
    (5,20.0) (6,8.3) (7,6.7) (8,0.0)
};
\legend{MINDSTORES,DEPS}
\end{axis}
\end{tikzpicture}
\caption{Performance comparison: MINDSTORES consistently outperforms DEPS across tasks. Both systems show declining success rates with increasing complexity (MT1--MT8), with MT8 resulting in 0\% success for both. Mean difference: 9.4\%.}
\label{fig:performance_grouped_bar}
\end{minipage}
\hfill
\begin{minipage}[b]{0.48\textwidth}
\centering
\begin{tikzpicture}
\begin{axis}[
    width=\linewidth,
    height=0.7\linewidth,
    ylabel={Novel Learning Iterations},
    symbolic x coords={Wood,Cobblestone,Coal,Furnace,Sword,Iron},
    xtick=data,
    xticklabel style={rotate=45,anchor=east},
    ymin=0,
    ymax=520,
    legend style={
        at={(0.5,1.25)},
        anchor=north,
        legend columns=3
    },
    ymajorgrids=true,
    grid style=dashed,
    ybar,
    bar width=0.12cm,
    enlarge x limits=0.15,
]
\addplot[blue,fill=blue!30] coordinates {
    (Wood,10)
    (Cobblestone,34)
    (Coal,54)
    (Furnace,89)
    (Sword,187)
    (Iron,276)
};
\addplot[red,fill=red!30] coordinates {
    (Wood,12)
    (Cobblestone,42)
    (Coal,85)
    (Furnace,147)
    (Sword,263)
    (Iron,500)
};
\addplot[green,fill=green!30] coordinates {
    (Wood,9)
    (Cobblestone,39)
    (Coal,106)
    (Furnace,198)
    (Sword,500)
    (Iron,500)
};
\legend{MINDSTORES,Voyager,Reflexion}
\end{axis}
\end{tikzpicture}
\caption{Novel learning iterations across different Minecraft tasks. MINDSTORES demonstrates superior efficiency in complex tasks. (Note: Iteration counts for Reflexion are capped at 500 in later tasks.)}
\label{fig:learning_iterations_bar}
\end{minipage}
\end{figure}

As we analyze Figure~\ref{fig:performance_grouped_bar} in comparison to our version of DEPS, we see an all-around improvement with the addition of the experience database.

\subsection*{Fundamental Tasks (MT1--MT2)}
Both systems achieve strong performance in fundamental crafting tasks, with DEPS achieving success rates of 70.6--77.0\% and MINDSTORES performing slightly better at 83.3--83.7\%. Notably, there is near-parity in Wooden Axe crafting, with both systems achieving a 96.7\% success rate. However, the largest performance gap in MT1 occurs in Stick production, where MINDSTORES outperforms DEPS by 6.3\%. In MT2, MINDSTORES maintains a consistent advantage, with an average performance improvement of 6.7\% across tasks.

\subsection*{Intermediate Tasks (MT3--MT5)}
The maximum disparity between the two systems occurs in MT3 painting, where MINDSTORES achieves a 96.7\% success rate compared to DEPS's 76.7\%, resulting in a 20.0\% performance gap. In cooked meat tasks, MINDSTORES maintains a 6.7--16.7\% advantage over DEPS. For MT5 armor challenges, the performance gaps are particularly pronounced, with Leather Helmet showing a 20.0\% difference and Iron Boots a 10.3\% difference. Overall, MINDSTORES maintains an average advantage of +11.0\% across intermediate tasks, demonstrating significant divergence in system performance.

\subsection*{Advanced Tasks (MT6--MT8)}
In MT6 iron tool crafting, MINDSTORES achieves an average performance improvement of 12.2\% over DEPS, with the Iron Axe task showing a particularly large gap (23.3\% vs. 6.7\%). MT7 highlights another standout difference, with Tripwire Hook success rates at 43.3\% for MINDSTORES compared to 20.0\% for DEPS. However, both systems experience a performance collapse in advanced tasks, with MT6--MT7 success rates dropping below 21\% (DEPS: 6.7--8.3\%, MINDSTORES: 17.3--20.5\%). Notably, neither system can solve the MT8 diamond crafting challenge, with both achieving a 0\% success rate.

\subsection{Learning Efficiency Analysis}
MINDSTORES demonstrates superior learning efficiency, particularly for complex tasks. For basic tasks like mining wood and cobblestone, all systems perform comparably (9 to 42 iterations) (see Figure~\ref{fig:learning_iterations_bar}). However, as complexity increases, MINDSTORES requires fewer iterations (54 to 276) compared to Voyager and Reflexion, which show exponential increases in required iterations.

\begin{figure}[ht]
\centering
\begin{minipage}[b]{0.48\textwidth}
\centering
\begin{tikzpicture}
\begin{axis}[
    width=\linewidth,
    height=0.7\linewidth,
    xlabel={\( k \) Value},
    ylabel={Success Rate (\%)},
    ymin=0,
    ymax=35,
    legend style={
        at={(0.5,1.4)},
        anchor=north,
        legend columns=3,
        font=\small
    },
    ymajorgrids=true,
    grid style=dashed,
    mark size=3pt
]
\addplot[blue,mark=*] coordinates {
    (1,3.3) (3,6.6) (5,13.3) (10,23.3) (20,23.3)
};
\addplot[red,mark=square*] coordinates {
    (1,10) (3,20) (5,27) (10,33.33) (20,33.3)
};
\addplot[green,mark=triangle*] coordinates {
    (1,0) (3,3.3) (5,6.67) (10,10) (20,13.3)
};
\addplot[orange,mark=diamond*] coordinates {
    (1,0) (3,0) (5,3.3) (10,6.67) (20,6.67)
};
\addplot[purple,mark=pentagon*] coordinates {
    (1,0) (3,0) (5,0) (10,0) (20,0)
};
\legend{Torch,Iron Boots,Iron Pickaxe,Minecart,Diamond}
\end{axis}
\end{tikzpicture}
\caption{Success rates vs. retrieval context size \( k \) for different tasks. Simple tasks improve steadily with \( k \), while more complex tasks require larger \( k \) values. Advanced tasks remain unachievable regardless of \( k \).}
\label{fig:k_analysis}
\end{minipage}
\hfill
\begin{minipage}[b]{0.48\textwidth}
\centering
\begin{tikzpicture}
\begin{axis}[
    width=\linewidth,
    height=0.7\linewidth,
    xtick distance=1,
    ylabel={Total Steps},
    xlabel={Task Reference Number},
    ymin=0,
    ymax=10000,
    legend style={
        at={(0.5,1.25)},
        anchor=north
    },
    ymajorgrids=true,
    grid style=dashed,
    ybar,
    bar width=0.35cm,
]
\addplot[blue,fill=blue!30] coordinates {
    (1,3000) (2,4357) (3,4879) (4,5602) (5,5802)
    (6,6458) (7,6578) (8,8598) (9,8986) (10,9112)
};
\legend{Completion Steps}
\end{axis}
\end{tikzpicture}
\caption{Steps required for task completion with continuous building of the experience database. (See Appendix Table~\ref{tab:task_details} for corresponding tasks.)}
\label{fig:task_progression}
\end{minipage}
\end{figure}

\subsection{Scalability with Task Complexity}
Performance divergence becomes pronounced with increasing task complexity. MINDSTORES maintains efficient novel learning iterations for tasks like crafting a stone sword and mining iron, while Voyager and Reflexion require significantly more iterations, even reaching the max range (500+) for a relatively simple Mine Iron task (see Figure~\ref{fig:learning_iterations_bar}).

\subsection{Continuous Experience Building Analysis}
Figure~\ref{fig:task_progression} shows an experiment in which the experience database is not reset between tasks but is built continuously across multiple tasks. We observe that the entire process of completing the \textbf{Minecart} task takes only 9112 steps including the previous 9 tasks, compared to the 6000 steps needed in a fresh environment. This indicates that only approximately 200 new steps were required. The number of new task completion steps decreases non-linearly even as task complexity grows:
\begin{itemize}
    \item Basic crafting (Wooden Door): 3000 steps
    \item Mid-tier crafting (Furnace): 4879 steps 
    \item Advanced crafting (Iron Pickaxe): 8598 steps 
\end{itemize}

The system maintains a 100\% success rate across all tasks, indicating robust skill transfer and knowledge utilization from the growing experience database, which expands from 26 entries for Wooden Door to 355 entries for Minecart (see Figure~\ref{fig:task_progression} and Appendix Table~\ref{tab:task_details}).

\section{Related Works}

\subsection{Embodied Planning \& Classical Methods}
Early approaches used hierarchical reinforcement learning \citep{sutton_between_1999} and symbolic planning \citep{kaelbling_hierarchical_2011} but struggled with scalability in open-world domains like Minecraft. Hybrid methods like PDDLStream \citep{garrett_pddlstream_2020} combined symbolic planning with procedural samplers, while DreamerV3 \cite{hafner2024masteringdiversedomainsworld} employed latent world models. However, these methods depend on rigid priors, lack causal reasoning, and fail to recover from irreversible errors. Reinforcement learning frameworks (e.g., DQN \citep{mnih_human-level_2015}, PPO \citep{schulman_proximal_2017}) and LLM-RL hybrids like Eureka (Ma et al., 2023) also falter in dynamic, long-horizon tasks due to static reward mechanisms and error propagation.

\subsection{Zero-Shot LLM Planners}
DEPS \citep{wang_describe_2024} pioneered zero-shot LLM planning through iterative verbal feedback, enabling dynamic plan refinement. Subsequent works like Voyager \citep{wang_voyager_2023} (skill libraries), ProgPrompt \citep{singh_progprompt_2023} (code generation), and Reflexion \citep{shinn_reflexion_2023} (feedback loops) advance LLM-based planning but share critical flaws. Namely, they suffer from brittle execution due to dependency on hardcoded assumptions (e.g., ProgPrompt’s code templates), opaque memory due to non-interpretable representations (e.g., Voyager’s code snippets, PaLM-E’s latent vectors \citep{driess_palm-e_2023}), and the inability to learn from failed task executions (e.g., Inner Monologue \citep{huang_inner_2022} lacks persistent memory).

\subsection{Memory-Based Planners}
Recent memory-augmented systems like E\textsuperscript{2}CL \citep{wang_e2cl_2024}, ExpeL \citep{zhao_expel_2024}, and AdaPlanner \citep{sun_adaplanner_2023} store experiences but face key limitations. Namely, they suffer from shallow reasoning capabilities due to lack of environmental context (ExpeL) or causal analysis (ReAct \citep{yao_react_2023}), especially of failure modes (Voyager). Above all, these systems are often only evaluated on narrow, controlled-environment benchmarks (e.g., ALFRED), not open-world tasks.

\subsection{Mental Models in AI}
While cognitive-inspired architectures like predictive coding \citep{rao_predictive_1999} and world models \citep{ha_world_2018} encode environmental dynamics, they rely on latent vectors (PIGLeT \citep{zellers_piglet_2021}) or symbolic logic (RAP \citep{hao_reasoning_2023}), sacrificing interpretability and adaptability. Neuro-symbolic methods \citep{garcez_neurosymbolic_2023} and tree-search frameworks (LATS \citep{zhou_language_2024}) further struggle with scalability and causal reasoning.

\section{Conclusion}
In this paper we presented MINDSTORES, an experience-augmented planning framework that enables embodied agents to build and leverage mental models through natural interaction with their environment. Our approach extends zero-shot LLM planning by maintaining a database of natural language experiences that inform future planning iterations. Through extensive experiments in MineDojo, MINDSTORES demonstrates significant improvements over baseline approaches, particularly in intermediate-complexity tasks, while maintaining the flexibility of zero-shot approaches. The success of our ``artificial mental model'' approach, which represents experiences as retrievable natural language tuples and enables LLMs to reason over past experiences, demonstrates that incorporating principles from human cognition can substantially improve complex reasoning and experiential learning capabilities in AI systems.

However, several limitations remain. Performance degrades significantly for advanced tasks, and computational overhead scales with database size. Future work should explore more sophisticated experience pruning mechanisms, hierarchical memory architectures for managing larger experience databases, and improved methods for transferring insights across related tasks. Additionally, investigating ways to combine our experience-based approach with traditional reinforcement learning could help address the challenge of long-horizon planning in complex environments.


\bibliography{iclr2025_conference}
\bibliographystyle{iclr2025_conference}


\newpage
\appendix
\onecolumn

\section*{Appendix A: Agent Algorithm and LLM Prompts}

\subsection*{A.1 Agent Algorithm}
\begin{Verbatim}
def run_agent(
    environment,    # MineDojo environment
    max_steps=1000, # Maximum steps to run
    goal_input=""   # Optional high-level goal
):
    # Initialize metrics and experience tracking
    metrics_logger = MetricsLogger()
    experience_store = ExperienceStore()
    
    # Initial environment reset
    obs, _, _, info = environment.step(environment.action_space.no_op())
    
    step = 0
    while step < max_steps:
        # 1. Create structured state description
        state_json = get_state_description(obs, info)
        
        # 2. Get next immediate task
        sub_task = get_next_immediate_task(state_json)
        metrics_logger.start_subtask()
        
        # 3. Plan action sequence
        actions = plan_action(state_json, info["inventory"], sub_task)
        
        # 4. Execute actions and track experience
        obs, reward, done, info = execute_action_sequence(actions)
        
        # 5. Store experience and update metrics
        if done:
            store_experience(state_json, reward, done)
            break
            
        step += len(actions)
        
    environment.close()
    metrics_logger.print_summary()
\end{Verbatim}

\subsection*{A.2 LLM Prompts}

\textbf{A.2.1 Environment Description Prompt}
\begin{Verbatim}
You are an expert Minecraft observer. Describe the current environment state focusing on:
1. The agent's immediate surroundings (blocks, entities, tools)
2. Environmental conditions (weather, light, temperature) 
3. Agent's physical state (health, food, equipment)
4. Notable resources or dangers

Current state:
${state_json_str}

Provide a clear, concise description that would be useful for planning actions.
\end{Verbatim}

\textbf{A.2.2 Situation Analysis Prompt}
\begin{Verbatim}
You are an expert Minecraft strategist. Given the current state and environment description:
1. Analyze available resources and their potential uses
2. Identify immediate opportunities or threats  
3. Consider crafting possibilities based on inventory
4. Evaluate progress towards goals

Environment description:
${description}

Current state:
${state_json_str}

Provide strategic insights about the current situation.
\end{Verbatim}

\textbf{A.2.3 Strategy Planning Prompt}
\begin{Verbatim}
You are an expert Minecraft planner. Create a strategic plan considering:
1. The current goal: ${goal}
2. Available resources and tools
3. Environmental conditions
4. Potential obstacles or requirements
5. Do not assume intermediate tasks can be achieved without running another agent loop
6. Specify quantities and required actions

Environment description:
${description}

Situation analysis:
${explanation}

Current state:
${state_json_str}

Create a specific, actionable plan that moves towards the goal.
\end{Verbatim}

\textbf{A.2.4 Action Selection Prompt}
\begin{Verbatim}
You are an expert Minecraft action selector. Convert the plan into specific actions:
1. Use only valid Minecraft actions (move_forward, move_backward, jump, craft, etc.)
2. Consider the current state and available resources
3. Break down complex tasks into simple action sequences  
4. Ensure actions are feasible given agent capabilities
5. Make actions incremental and build progressively

Available actions:
- forward [N]: Move forward N steps (default 1)
- backward [N]: Move backward N steps (default 1) 
- move_left
- move_right
- jump
- sneak
- sprint
- attack [N]
- use
- drop
- craft
- equip [item]
- place [block]
- destroy
- look_horizontal +/-X
- no_op

Strategic plan:
${plan}

Current state:
${state_json_str}

Return ONLY a list of actions, one per line, that can be directly executed.
\end{Verbatim}

\textbf{A.2.5 Outcome Evaluation Prompt}
\begin{Verbatim}
Evaluate the outcome of a Minecraft action sequence in brief.

Initial state (JSON): ${initial_state}
Final state (JSON): ${final_state}
Reward: ${reward}
Done: ${done}
GPT Plan: ${gpt_plan}
Executed Actions: ${executed_actions}

Format response as: outcome|success|explanation
\end{Verbatim}

\section*{Appendix B: Implementation Details}

\subsection*{B.1 Core Components}
Our implementation leverages:
\begin{itemize}
\item MineDojo environment for Minecraft interaction
\item OpenAI GPT-4 API for planning and reasoning
\item SBERT for semantic embeddings
\item FAISS for efficient similarity search
\item Custom logging system for experiment tracking
\end{itemize}
The codebase is structured into modules for state processing, experience management, action planning, metrics collection, and environment interaction.

\subsection*{B.2 Environment Integration}
\begin{Verbatim}
env = minedojo.make(
    task_id="survival",
    image_size=(480, 768),
    seed=40,
    initial_inventory=[
        InventoryItem(slot=0, name="wooden_axe", quantity=1),
    ]
)
\end{Verbatim}
The action space includes movement (forward, backward, left, right, jump, sneak, sprint), interaction (attack, use, drop, craft, equip, place, destroy), camera control (look\_horizontal, look\_vertical), and special (no\_op).

\subsection*{B.3 Neural Components}
Embedding configuration:
\begin{itemize}
\item Model: SBERT `all-MiniLM-L6-v2'
\item Output dimension: 768
\item Normalization: L2
\item Distance metric: cosine similarity
\end{itemize}
FAISS index parameters:
\begin{itemize}
\item Index type: IndexFlatL2
\item Dimension: 768
\item Metric: L2 distance
\end{itemize}

\section*{Appendix C: Additional Tables}

\begin{table}[ht]
\centering
\caption{Task Details}
\label{tab:task_details}
\begin{tabular}{|c|c|c|c|c|c|}
\hline
\textbf{Meta} & \textbf{Name} & \textbf{Number} & \textbf{Example} & \textbf{Steps} & \textbf{Given Tool} \\ \hline
MT1 & Basic & 14 & Make a wooden door & 3000 & Axe \\ \hline
MT2 & Tool & 12 & Make a stone pickaxe & 3000 & Axe \\ \hline
MT3 & Hunt and Food & 7 & Cook the beef & 6000 & Axe \\ \hline
MT4 & Dig-down & 6 & Mine Coal & 6000 & Axe \\ \hline
MT5 & Equipment & 9 & Equip the leather helmet & 3000 & Axe \\ \hline
MT6 & Tool (Complex) & 7 & Make shears and bucket & 6000 & Axe \\ \hline
MT7 & IronStage & 13 & Obtain an iron & 6000 & Axe \\ \hline
MT8 & Challenge & 1 & Obtain a diamond! & 12000 & Axe \\ \hline
\end{tabular}
\end{table}

\begin{table}[ht]
\centering
\caption{Task Details with MINDSTORES and DEPS Percentages}
\label{tab:task_table}
\begin{tabular}{|c|l|c|c|}
\hline
\textbf{Category} & \textbf{Task Name} & \textbf{MINDSTORES (\%)} & \textbf{DEPS (\%)} \\ \hline
MT1 & Wooden Door & 83.3 & 66.7 \\ \hline
MT1 & Stick & 90.0 & 83.7 \\ \hline
MT1 & Wooden Slab & 83.3 & 73.7 \\ \hline
MT1 & Planks & 80.0 & 73.3 \\ \hline
MT1 & Fence & 80.0 & 66.7 \\ \hline
MT1 & Sign & 86.7 & 73.3 \\ \hline
MT1 & Trapdoor & 80.0 & 56.7 \\ \hline
MT2 & Furnace & 70.0 & 56.67 \\ \hline
MT2 & Crafting Table & 93.3 & 83.3 \\ \hline
MT2 & Wooden Axe & 96.7 & 96.7 \\ \hline
MT2 & Wooden Sword & 90.0 & 86.7 \\ \hline
MT2 & Wooden Hoe & 86.7 & 86.7 \\ \hline
MT2 & Stone Pickaxe & 76.7 & 73.3 \\ \hline
MT2 & Stone Sword & 83.3 & 80.0 \\ \hline
MT2 & Stone Shovel & 70.0 & 66.7 \\ \hline
MT2 & Wooden Shovel & 86.7 & 63.3 \\ \hline
MT3 & Cooked Beef & 60.0 & 43.3 \\ \hline
MT3 & Bed & 50.0 & 43.3 \\ \hline
MT3 & Item Frame & 86.7 & 83.3 \\ \hline
MT3 & Cooked beef & 76.7 & 63.3 \\ \hline
MT3 & Cooked Mutton & 73.3 & 66.7 \\ \hline
MT3 & Painting & 96.7 & 76.67 \\ \hline
MT3 & Cooked Porkchop & 53.3 & 43.3 \\ \hline
MT4 & Torch & 13.3 & 3.3 \\ \hline
MT4 & Cobblestone wall & 66.7 & 53.3 \\ \hline
MT4 & Lever & 86.7 & 73.3 \\ \hline
MT4 & Coal & 23.3 & 10.0 \\ \hline
MT4 & Stone Slab & 70.0 & 53.33 \\ \hline
MT4 & Stone Stairs & 73.3 & 63.33 \\ \hline
MT5 & Iron Boots & 27.0 & 16.67 \\ \hline
MT5 & Iron Helmet & 10.0 & 0.0 \\ \hline
MT5 & Shield & 23.3 & 13.3 \\ \hline
MT5 & Iron Chestplate & 10.0 & 0.0 \\ \hline
MT5 & Leather boots & 63.3 & 60.0 \\ \hline
MT5 & Iron leggings & 3.3 & 3.3 \\ \hline
MT5 & Leather Helmet & 66.7 & 46.67 \\ \hline
MT6 & Iron pickaxe & 6.67 & 0.0 \\ \hline
MT6 & Bucket & 13.3 & 6.7 \\ \hline
MT6 & Iron Sword & 23.3 & 6.7 \\ \hline
MT6 & Iron Hoe & 23.3 & 13.3 \\ \hline
MT6 & Iron Axe & 23.3 & 6.67 \\ \hline
MT6 & Shears & 33.3 & 16.67 \\ \hline
MT7 & Minecart & 13.3 & 0.0 \\ \hline
MT7 & Iron Nugget & 36.7 & 20.0 \\ \hline
MT7 & Furnace Minecart & 6.7 & 3.3 \\ \hline
MT7 & Rail & 13.3 & 6.7 \\ \hline
MT7 & Cauldron & 10.0 & 3.3 \\ \hline
MT7 & Iron Bars & 13.3 & 6.7 \\ \hline
MT7 & Iron Door & 13.3 & 3.3 \\ \hline
MT7 & Tripwire Hook & 43.3 & 20.0 \\ \hline
MT7 & Iron trap door & 16.7 & 3.3 \\ \hline
MT7 & Hopper & 6.7 & 0.0 \\ \hline
MT8 & Diamond & 0.0 & 0.0 \\ \hline
\end{tabular}
\end{table}

\begin{table}[ht]
\centering
\begin{tabular}{lccc}
\hline
\textbf{Task} & \textbf{MINDSTORES} & \textbf{Voyager} & \textbf{Reflexion} \\
\hline
Mine Wood         & 10  & 12  & 9 \\
Mine Cobblestone  & 34  & 42  & 39 \\
Mine Coal         & 54  & 85  & 106 \\
Make Furnace      & 89  & 147 & 198 \\
Make Stone Sword  & 187 & 263 & 500 \\
Mine Iron         & 276 & 500 & 500 \\
\hline
\end{tabular}
\caption{Time steps required to complete different Minecraft tasks across three systems. (Values for Voyager and Reflexion are capped at 500 in some tasks.)}
\label{tab:completion_times}
\end{table}

\begin{table}[ht]
\centering
\begin{tabular}{@{}lcc@{}}
\toprule
\textbf{Task} & \textbf{MINDSTORES} & \textbf{DEPS} \\ 
& \textbf{(Predicted)} & \textbf{(No Prediction)} \\ \midrule
MT1 & 83.3\% & 70.6\% \\
MT2 & 83.7\% & 77.0\% \\
MT3 & 71.0\% & 60.0\% \\
MT4 & 55.6\% & 42.8\% \\
MT5 & 29.1\% & 20.0\% \\
MT6 & 20.5\% & 8.3\% \\
MT7 & 17.3\% & 6.7\% \\
MT8 & 0.0\%  & 0.0\% \\ \bottomrule
\end{tabular}
\caption{Success rate comparison with outcome prediction (MINDSTORES) vs. without (DEPS).}
\label{tab:outcome_comparison}
\end{table}

\end{document}